\documentclass[10pt,twocolumn,letterpaper]{article}

\usepackage{iccv}
\usepackage{times}
\usepackage{epsfig}
\usepackage{graphicx}
\usepackage{amsmath}
\usepackage{amssymb}

\usepackage{booktabs}
\usepackage{inconsolata}
\usepackage{multirow}
\usepackage{pifont}

\newcommand{\cmark}{\ding{51}}%
\newcommand{\xmark}{\ding{55}}%

\usepackage[pagebackref=true,breaklinks=true,letterpaper=true,colorlinks,bookmarks=false]{hyperref}

\iccvfinalcopy 

\def\iccvPaperID{1105} 

\ificcvfinal\fi

\begin{document}

\title{\vspace{-0.35cm} MAAS: Multi-modal Assignation for Active Speaker Detection \vspace{-0.35cm}}

\author{
    Juan León Alcázar$^{1}$, Fabian Caba Heilbron$^{2}$, Ali K. Thabet$^{1}$ \& Bernard Ghanem$^{1}$\\
    \small $^{1}$ King Abdullah University of Science and Technology (KAUST), $^{2}$Adobe Research\\
    {\tt\small jc.leon@uniandes.edu.co, caba@adobe.com, ali.thabet@kaust.edu.sa, bernard.ghanem@kaust.edu.sa}
}

\maketitle
\ificcvfinal\thispagestyle{empty}\fi

\begin{abstract}
    Active speaker detection requires a mindful integration of multi-modal cues. Current methods focus on modeling and fusing short-term audiovisual features for individual speakers, often at frame level. We present a novel approach to active speaker detection that directly addresses the multi-modal nature of the problem and provides a straightforward strategy, where independent visual features (speakers) in the scene are assigned to a previously detected speech event. Our experiments show that a small graph data structure built from local information can approximate an instantaneous audio-visual assignment problem. Moreover, the temporal extension of this initial graph achieves a new state-of-the-art performance on the AVA-ActiveSpeaker dataset with a mAP of 88.8\%. We have made our code available at \url{https://github.com/fuankarion/MAAS}.
\end{abstract}


\section{Introduction}
\label{sec:intro}

Active speaker detection aims at identifying the current speaker (if any) from a set of candidate face detections in an arbitrary video. This research problem is an inherently multi-modal task that requires the integration of subtle facial motion patterns and the characteristic waveform of speech. Despite its multiple applications such as speaker diarization \cite{anguera2012speaker,  shum2013unsupervised, tranter2006overview, wang2018speaker}, human-computer interaction \cite{garcia2017speaker, yu2017active} and bio-metrics \cite{nagrani2018seeing, ravanelli2018speaker}, the detection of active speakers in-the-wild remains an open problem. 

Current approaches for active speaker detection are based on recurrent neural networks \cite{roth2019ava, sharma2020crossmodal} or 3D convolutional models \cite{afouras2020self, chung2019naver, zhangmulti}. Their main focus is to jointly model the audio and visual streams to maximize the performance of single speaker prediction over short sequences. Such an approach is suitable for single speaker scenarios, but is overly simplified for the general (multi-speaker) case.

\begin{figure}[t!]
    \begin{center}
        \includegraphics[width=0.5\textwidth]{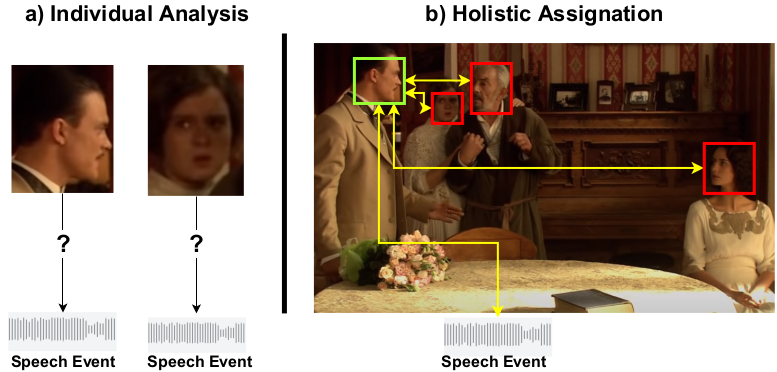}
    \end{center}
    \caption{\textbf{Audiovisual assignment for active speaker detection.} Active speaker detection is highly ambiguous. Even if we analyze joint audiovisual information, unrelated facial gestures can easily resemble the natural motion of lips while speaking. In \textbf{a)} we show two face crops from a sequence, where a speech event was detected. The gestures, illumination, and capture angle make it hard to asses which face (if any) is the active speaker. Our strategy \textbf{b)} focuses on the attribution of speech segments in video. If a speech event is detected, we holistically analyse every speaker along with the audio track to discover the most likely active speaker.
    }
    \label{fig:teaser}
\end{figure}

The general (multi-speaker) scenario has two major challenges. First, the presence of multiple speakers allows for incorrect face-voice assignations. For instance, false positives emerge when facial gestures closely resemble the motion patterns observed while speaking (\eg laughing, grinning). Second, it must enforce temporal consistency over multi-modal data, which quickly evolves over time, \eg, when active speakers switch during a fluid conversation. 

In this paper, we address the general multi-speaker problem in a principled manner. Our key insight is that, instead of optimizing active speaker predictions over individual audiovisual embeddings, we can jointly model a set of visual representations from every speaker in the scene along with a single audio representation extracted from the shared audio track. While simple, this modification allows us to map the active speaker detection task into an assignation problem, whose goal is to match multiple visual representations with a singleton audio embedding. Figure \ref{fig:teaser} illustrates some of the challenges in active speaker detection and provides a general insight for our approach. 
 
Our approach, dubbed ``Multi-modal Assignation for Active Speaker detection'' (MAAS) relies on multi-modal graph neural networks \cite{kipf2016semi, wang2018dynamic} to approach the local (frame-wise) assignation problem, but it is flexible enough to also propagate information from a long-term analysis window by simply updating the underlying graph connectivity. In this framework, we define the active speaker as the local visual representation with the highest affinity to the audio embedding. Our empirical findings highlight that reformulating the problem into a multi-modal assignation problem brings sizable improvements over current state-of-the-art methods. On the AVA Active speaker benchmark, MAAS outperforms all other methods by at least 1.7\%. Additionally, when compared with methods that analyze a short temporal span, MAAS brings a performance boost of at least 1.1\%.

\noindent\textbf{Contributions.} This paper proposes a novel strategy for active speaker detection, which explicitly learns multi-modal relationships between audio and facial gestures by sharing information across modalities. Our work brings the following contributions: \textbf{(1)} We devise a novel formulation for the active speaker detection problem. It explicitly matches the visual features from multiple speakers to a shared audio embedding of the scene (Section \ref{subsec:LAN}). 
\textbf{(2)} We empirically show that this assignation problem can be solved by means of a Graph Convolutional Network (GCN), which endows flexibility on the graph structure and is able to achieve state of the art results (Section \ref{subsec:sota}). 
\textbf{(3)} We present a novel dataset for active speaker detection, called ``Talkies", as a new benchmark composed of 10,000 short clips gathered from challenging and diverse scenes (Section \ref{sec:talkies}).

To ensure reproducible results and promote future research, all the resources of this project, including source code, model weights, official benchmark results, and labeled data will be publicly available.

\section{Related Work}

\begin{figure*}[t!]
    \begin{center}
        \includegraphics[width=1\textwidth]{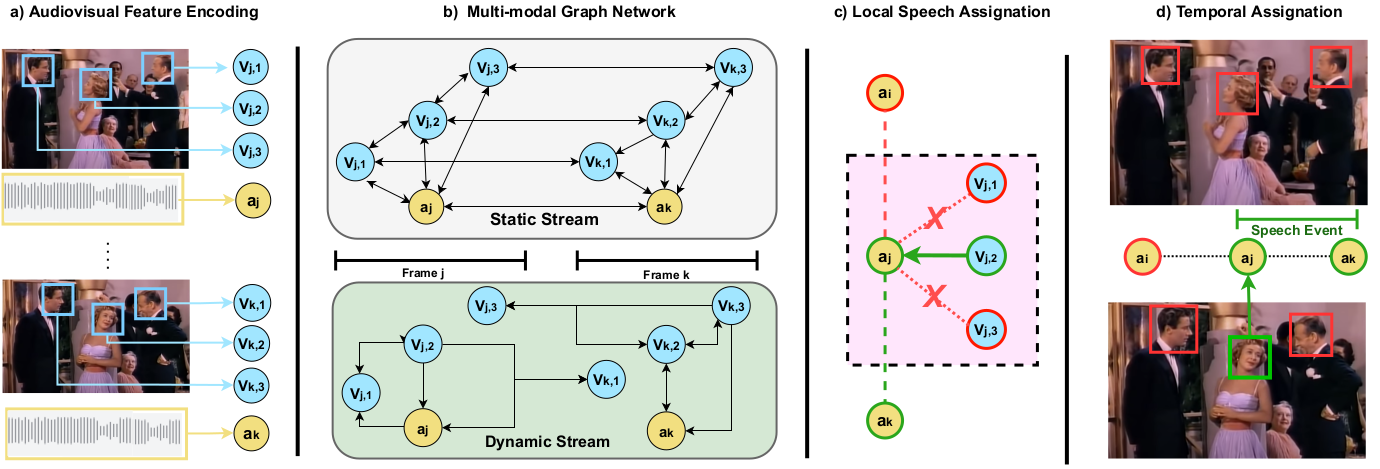}
    \end{center}
    \caption{\textbf{Overview of MAAS Pipeline.} 
    \textbf{a)}: Our approach begins by sampling independent audio and video features. Video features (cyan) are extracted from a stack of face crops that belong to a single person. Audio features (yellow) are extracted from the audio-spectrogram and are shared at the frame-level. \textbf{b)}: We create two feature graphs: one with static connections that model local temporal relations between the audio track and the visible persons; in parallel, we allow a secondary stream in the network to discover relationships given the estimated feature embeddings. \textbf{c)}: We estimate a frame-level affinity between the visual nodes and the local audio node such that the active speaker will have the highest affinity with the audio node. \textbf{d)}: Finally, we extend the network by modelling a longer temporal window. We jointly optimize the local affinities, while enforcing temporal consistency. We select the active speaker (green bounding box) as the most likely speaker to have generated the sequence of speech events. 
    }
    \label{fig:Overview}
\end{figure*}

In the realm of multi-modal learning, different information sources are fused with the goal of establishing more effective representations \cite{ngiam2011multimodal}. In the video domain, a common multi-modal paradigm involves combining representations from visual and audio features \cite{chakravarty2015s, chung2018voxceleb2, jati2019neural, nagrani2018learnable, nagrani2018seeing, owens2018audio, wang2018voicefilter}. Such representation allows the exploration of new approaches to well established problems, such as person re-identification \cite{nagrani2018learnable, kim2018learning, yadav2018learning}, audio-visual synchronization \cite{afouras2020self, chung2017lip, chung2016out}, speaker diarization \cite{shum2013unsupervised, wang2018speaker, zhang2019fully}, bio-metrics \cite{nagrani2018seeing, ravanelli2018speaker}, and audio-visual source separation \cite{chakravarty2015s, jati2019neural, owens2018audio, roth2019ava, wang2018voicefilter}. Active speaker detection is a special instance of audiovisual source separation, where the sources are people in a video, and the goal is to detect and assign a segment of speech to one of those sources \cite{roth2019ava}.

\noindent
\textbf{Active Speaker Detection.}
The work of Cutler \etal \cite{cutler2000look} pioneered research in active speaker detection in the early 2000s. It detected correlated audiovisual signals by means of a time-delayed neural network \cite{waibel1989phoneme}. Follow up works \cite{everingham2009taking, saenko2005visualspeech}  approached the task relying only on visual information, focusing strictly on the evolution of facial gestures. Such visual-only modeling was possible as they addressed a simplified version of the problem with a single candidate speaker. Recent works \cite{chakravarty2016active, chung2016out} have approached the more general multi-speaker scenario and relied on fusing multi-modal information from individual speakers. A parallel corpus of work has focused on audiovisual feature alignment, which resulted in methods that rely on audio as the primary source of supervision \cite{chakravarty2015s}, or as an alternative to jointly train a deep audiovisual embedding \cite{chung2018voxceleb2, chung2016out, nagrani2017voxceleb, tao2017bimodal}.

The work of Roth \etal \cite{roth2019ava} introduced the AVA-ActiveSpeaker dataset and benchmark, the first large-scale video dataset for the active speaker detection task. In the AVA-ActiveSpeaker challenge of 2019, Chung \etal \cite{chung2019naver} presented an improved architecture of their previous work \cite{chung2016out}, which trains a large 3D model with the need for large-scale audiovisual pre-training \cite{nagrani2017voxceleb}. Zhang \etal \cite{zhangmulti} also leveraged a hybrid 3D-2D architecture with large-scale pre-training \cite{chung2016out, chung2019perfect}. This method achieved its best performance when the feature embedding was optimized using a contrastive loss \cite{hadsell2006dimensionality}. Follow up works focused on modeling an attention process over face tracks, where attention was estimated either from the audio alignment \cite{afouras2020self} or from an ensemble of speaker features \cite{alcazar2020active}. We approach the active speaker problem in a more principled manner, as we go beyond the aggregation of contextual information from multiple-speakers and propose an approach that explicitly seeks to model the correspondence of a shared audio embedding with all potential speakers in the video.

\noindent
\textbf{Datasets for Active Speaker Detection.}
Apart from the development of the AVA-ActiveSpeaker benchmark, there are few public datasets specific to this problem. The most well known alternative is the Columbia dataset \cite{chakravarty2016active}, which contains 87 minutes of labeled speech from a panel discussion. It is much smaller and less diverse than AVA. Modern audiovisual datasets \cite{nagrani2017voxceleb, chung2016out, chung2020spot} have been adapted for the large scale pre-training of some active speaker methods \cite{chung2019naver}. Nevertheless, these datasets were designed for related tasks such as speaker identification and speaker diarization. In this paper, we present the Talkies dataset as a new benchmark for active speaker detection gathered from social media clips. It contains 800,000 manually labeled face detections and includes challenging scenarios that contain multiple speakers, occlusion, and out of screen speech.

\noindent
\textbf{Graph Convolutional Networks (GCNs).}
GCNs~\cite{kipf2016semi} have recently gained popularity, due to the greater interest in non-Euclidean data. In computer vision, GCNs have been successfully applied to scene graph generation \cite{cv_inv_scene_johnson2018image, cv_scene_li2018factorizable, qi20173d, cv_scene_xu2017scene, cv_scene_yang2018graph}, 3D understanding \cite{Gkioxari2019Mesh, li2019sgas, wang2018dynamic, xie2019clouds}, and action recognition in video \cite{cv_action_jain2016structural, g_tad, cv_action_yan2018spatial}. In MAAS, we design a DeepGCN-like architecture \cite{Li_2019_ICCV,li2019deepgcns_journal,li2020deepergcn}, which addresses a special scenario, namely the multi-modal nature of audiovisual data. We rely on the well-known EdgeConv operator \cite{wang2018dynamic} to model interactions between different modalities for graph nodes identified across multiple frames. This enables us to model both the multi-modal relations and the temporal dependencies in a single graph structure.

\section{Multi-modal Active Speaker Assignation}

Our approach is based on a straightforward idea. Instead of assessing the likelihood of individual audiovisual patterns to belong to an active speaker, we directly model the correspondence between the local audio and the facial gestures of all the individuals present in the scene. This approach is motivated by the nature of the active speaker problem, which first identifies if any speech patterns are present, and then attributes those patterns to a single speaker.

Overall, our approach simultaneously solves three sub-tasks. First, we detect speech events in a short-term temporal window. Second, we iterate over all the visible speakers in a single frame, and decide which one is most likely to be an active speaker given the local information. Third, we extend this frame-level analysis along the temporal dimension, leveraging the inherent temporal consistency of video data to improve frame-level predictions. Figure \ref{fig:Overview} illustrates an overview of our MAAS approach.

\subsection{Frame-Level Video Features}
\label{subsec:Features}

Following recent works \cite{roth2019ava, zhangmulti}, we extract the initial frame-level features from a two-stream convolutional encoder. The visual stream takes as input a tensor of dimensions ${H \times W \times (3 c)}$, where $H$ and $W$ are the image width and height, and $c$ is the number of time consecutive face crops sampled from a single tracklet. Similar to \cite{roth2019ava}, we transform the original audio waveform into a Mel-spectrogram and use it as input for the audio stream. 

Our approach relies on independent audio and video features. To obtain these independent features (and to make fair comparison to state-of-the-art techniques), we train a joint model as described by \cite{roth2019ava}, but drop the final two layers at inference time. These layers are responsible for the feature fusion and final prediction. 

At time $t$, a forward pass of our feature encoder yields $N+1$ feature vectors for a frame with $N$ possible speakers (detected persons). One shared audio embedding ($\mathbf{a}_{t}$) and $N$ independent visual descriptors $\mathbf{v}_{t} = \{ v_{t,0}, v_{t,1}, ..., v_{t,n-1} \}$ one for each of the $N$ visible persons (see Figure \ref{fig:Overview}-a). We define ($\mathbf{s_{t}}$) as the local set of features at time $t$, such that $\mathbf{s_{t}} =\{\mathbf{a}_{t} \cup \mathbf{v}_{t} \}$. The feature set $\mathbf{s_{t}}$ is used for the optimization of the basic graph structure in MAAS, the Local Assignation Network, described next.

\begin{figure*}[t!]
    \begin{center}
        \includegraphics[width=0.98\textwidth]{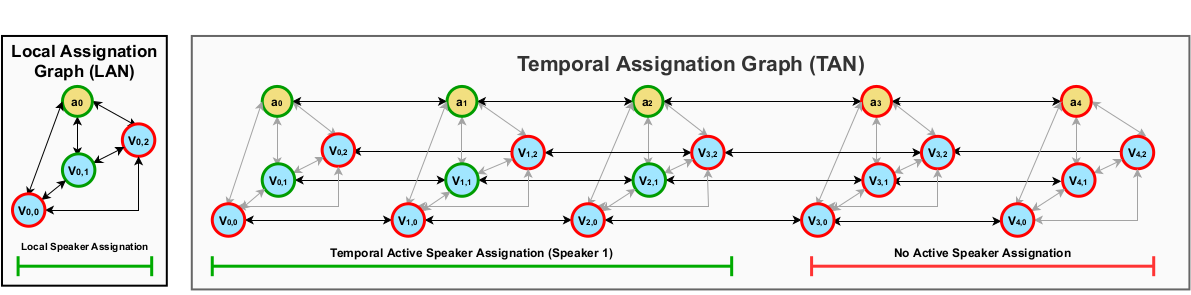}
    \end{center}
    \caption{\textbf{Assignation Graphs.} The base static graphs for MAAS are composed of multi-modal nodes, visual nodes (Cyan), and audio nodes  (Yellow). The local Assignation Graph (left) defines frame level connectivity of the individual features. The Temporal Assignation Graph (right) is composed of multiple local graphs (5 in the figure) and defines a temporal extension of the frame-level relations (we depict local relations in light gray to avoid visual clutter). While a local graph solves an instantaneous assignation problem.  The temporal graph optimizes a subset of nodes thereby incorporating temporal information in the individual local graphs.
    }
    \label{fig:tempralgraph}
\end{figure*}

\subsection{Local Assignation Network (LAN)}
\label{subsec:LAN}

We model the local assignation problem by generating a directed graph over the feature set $\mathbf{s_{t}}$. Our local graph consists of an audio node and one video node for each potential speaker. We create a bidirectional connectivity between the audio node and each visual node thus leveraging a GCN that operates on a directed graph generated from $\mathbf{s_{t}}$. Figure \ref{fig:tempralgraph} (left) illustrates this graph structure. We call this graph structure the Local Assignation Graph and the GCN that operates over it the Local Assignation Network (LAN).

The goals of LAN are two-fold: (i) to detect local speech events, (ii) if there is a speech event, to assign the most likely speaker from the set of candidates. We achieve these two goals by fully supervising every node in LAN. Visual nodes are supervised by the ground-truth, $l_{tv}$, of the corresponding speaker. On the other hand, each audio node receives a binary ground-truth label indicating whether there is at least one active speaker, \ie $\max(\{l_{t,0}, l_{t,1}, ... , l_{t,n-1}\})$; otherwise, there is silence. LAN is tasked to discover active speakers at the frame-level (\ie $t$ is fixed).

\subsection{Temporal Assignation Network (TAN)}
\label{subsec:TAN}

While the LAN is effective at finding local correspondences between audio patterns and visible faces, it models information sampled from short video clips ($\mathbf{s_{t}}$). This sampling strategy can lead to inaccurate predictions from noisy or ambiguous local estimations (\eg audio noise, blurred faces, ambiguous facial gestures, etc.). Therefore, we extend our proposed approach to include temporal information from adjacent frames.

We extend the local graph in LAN by sampling $\mathbf{s_{t}}$ over a temporal window ($w$) centered around time $t$.  $w = [i, i+1, ..., t, ..., j]$ and define a temporal feature set $\mathbf{b}_{w} = [\mathbf{s}_{i}, \mathbf{s}_{i+1},..., \mathbf{s}_{t}, ...,  \mathbf{s}_{j}]$. Following the LAN structure outlined in \ref{subsec:LAN}, we can build $(j-i)$ independent local graph structures out of $\mathbf{b}_{w}$ (one for every time step). We augment this set of independent graphs by adding temporal links between time adjacent representations of frame-level features. We follow two rules to build these connections: we create temporal connections between time adjacent audio nodes, and we create temporal connections between time adjacent video nodes, only if they belong to the same person. No additional cross-modal connections are built. We call the resulting graph, the Temporal Assignation Graph, which allows for information flow between time adjacent audio and video features, thereby allowing for temporal consistency in the audio and video modalities. Figure \ref{fig:tempralgraph} (right) illustrates this graph structure. 

We build a GCN over the extended graph topology and call it the Temporal Assignment Network (TAN). TAN allows us to directly identify speech segments as continuous positive predictions over audio nodes. Likewise, it detects active speech segments over continuous predictions for same-speaker video features.

\subsection{Dynamic Stream \& Global Prediction}

Finally, we account for potential connection patterns that go beyond our initial insights. We augment our architecture and define a second stream that will operate on the very same data as the static stream (including multiple temporal timestamps). However, we do not define a fixed connectivity pattern for this stream. Instead, we aim at creating a dynamic graph structure based on the node distribution in feature space. In this stream, we allow the GCN to estimate an arbitrary graph structure by calculating the $K$ nearest neighbors in feature space for each node, and by establishing edges based on these neighbouring nodes. In practice, we replicate the static stream, drop the definition of the static graph, and use the dynamic version of the edge-convolution \cite{wang2018dynamic}, allowing for independent dynamic graph estimation at every layer.

\begin{figure}[t]
    \begin{center}
        \includegraphics[width=0.45\textwidth]{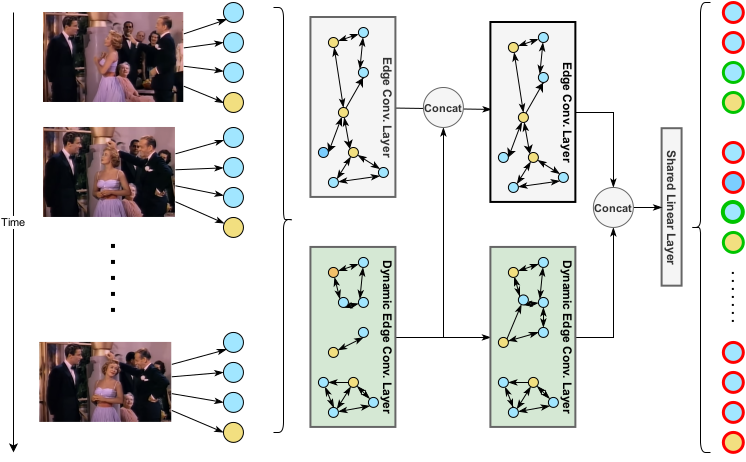}
    \end{center}
    \caption{\textbf{GCN Architecture in MAAS.}
    Our graph neural network implements a two stream architecture. The first stream (top) uses the edge convolution operator and operates over static local and temporal graphs. The second stream (bottom) relies on dynamic edge convolutions and complements the feature embedding discovered by the static stream by means of slow fusion. After every GCN layer, we fuse the features from the dynamic and static streams and use them as input to the next layer.
    }
    \label{fig:ourgcn}
\end{figure}

The final prediction is achieved through slow fusion \cite{karpathy2014large, g_tad}. At every GCN layer, we merge the feature set from the dynamic layer with the feature set from the static layers. The final prediction is achieved using a shared fully connected layer and softmax activation over every node. This architecture is depicted in Figure \ref{fig:ourgcn}.

\subsection{Training and Implementation Details}

Following \cite{roth2019ava}, we implement a two-stream feature encoder based on the ResNet-18 architecture \cite{he2016deep} pre-trained on ImageNet \cite{deng2009imagenet}. We perform the same modifications at the first layer to adapt for the extended input tensor (stack of face crops and spectrogram). We train the network end-to-end using the Pytorch library \cite{paszke2017automatic} for 100 epochs with the ADAM optimizer \cite{kingma2014adam} using Cross-Entropy Loss.  We use $3\times 10^{-4}$ as initial learning rate that decreases with annealing of $\gamma =0.1$ at epochs 40 and 80. We empirically set $c=11$ and augment the input videos via random flips and corner crops. Unlike other methods, MAAS does not require any large-scale audiovisual pre-training. We also incorporate the sampling strategy proposed by \cite{alcazar2020active} in training to alleviate  overfitting. During training, we follow the supervision strategy outlined by \cite{roth2019ava}, where two extra auxiliary loss functions $ (\mathcal{L}_{a} , \mathcal{L}_{v})$ are adopted to supervise the final layer of the audio and video streams. This favors the estimation of useful features from both streams. 

\paragraph{Training MAAS}
After optimizing the feature encoder, we implement MAAS (LAN and TAN networks) using the PyTorch Geometric library \cite{Fey/Lenssen/2019}. We choose edge-convolution \cite{wang2019dynamic} to propagate the neighbor information between nodes. Our network model contains 4 GCN layers on both streams, each with filters of 64 dimensions. We apply dimensionality reduction to map features from their original $512$ dimension to $64$ using a fully connected layer. We find that this dimensionality reduction favors the final performance and largely reduces the computational cost.

Since we process data from different modalities, we use two different dimensionality reduction layers, one for video features and another for audio features.  We train the MAAS-LAN and MAAS-TAN networks using the same procedure and set of hyper-parameters, the only difference being their underlying graph structure. We use the ADAM optimizer with an initial learning rate of $3\times 10^{-4}$ and train for 4 epochs. Both GCNs are trained from random weights and use a pre-activation \cite{he2016identity} linear layer (Batch Normalization $\rightarrow$ ReLU $\rightarrow$ Linear Layer) to map the concatenated node features inside the edge convolution.

\section{Experimental Validation}
\label{sec:results}

In this section, we provide an empirical analysis of our proposed MAAS method. We focus on the large-scale AVA-ActiveSpeaker dataset \cite{roth2019ava} to assess the performance of MAAS and present additional evaluation results on Talkies. This section is divided in three parts. First, we compare MAAS with state-of-the-art techniques. Then, we ablate our proposal to analyse all of its individual design choices. Finally, we test MAAS on known challenging scenarios to explore common failure modes.

\paragraph{AVA-ActiveSpeaker Dataset.} The AVA-ActiveSpeaker dataset \cite{roth2019ava} is the first large-scale testbed for active speaker detection. It is composed of $262$ Hollywood movies: $120$ of those on the training set, $33$ on validation, and the remaining $109$ on testing. The AVA-ActiveSpeaker dataset contains normalized bounding boxes for 5.3 million faces, all manually curated from automatic detections. Facial detections are manually linked across time to produce face tracks (tracklets) depicting a single identity. Each face detection is labeled as speaking, speaking but not audible, or non-speaking. All AVA-ActiveSpeaker results reported in this paper were measured using  the official evaluation tool provided by the dataset creators, which uses mean average precision (mAP) as the main metric for evaluation.

\subsection{State-of-the-art Comparison}
\label{subsec:sota}

\begin{table}[t!]
    \small
    \centering
    \begin{tabular}{ l c }
    \toprule
    \textbf{Method}  & \textbf{mAP} \\
    \midrule
    \multicolumn{2}{l}{\textit{Validation Set}} \\
    \textbf{MAAS-TAN (Ours)} &  \textbf{88.8} \\
    Alcazar \etal \cite{alcazar2020active} &  87.1  \\
    Chung \etal (Temporal Convolutions)  \cite{chung2019naver} & 85.5 \\
    Alcazar \etal (Temporal Context) \cite{alcazar2020active}   &   85.1 \\
    Chung \etal (LSTM) \cite{chung2019naver}   &   85.1 \\
    \textbf{MAAS-LAN (Ours)} &  85.1 \\
    Zhang \etal \cite{zhangmulti} & 84.0 \\
    Sharma \etal \cite{sharma2020crossmodal} & 82.0 \\
    Roth \etal  \cite{roth2019ava} & 79.2 \\

    \midrule
    \midrule
    \multicolumn{2}{l}{\textit{Test Set}} \\
    \textbf{MAAS-TAN (Ours)} & \textbf{88.3} \\
    Naver Corporation \cite{chung2019naver} & 87.8 \\
    Active Speaker Context \cite{alcazar2020active} & 86.7  \\ 
    University of Chinese Academy of Sciences \cite{zhangmulti} & 83.5 \\
    Google Baseline \cite{roth2019ava} & 82.1 \\

    \toprule
    
    \end{tabular}
    \caption{\textbf{State-of-the-art Comparison on AVA-ActiveSpeaker.} 
    We compare MAAS against state-of-the-art methods on the AVA-ActiveSpeaker validation set. Results are measured with the official evaluation tool as published by \cite{roth2019ava}. We report an improvement of 1.7\% mAP over the current state-of-the-art.
    }
    \label{tab:sota}
\end{table}

We begin our analysis by comparing MAAS to state-of-the-art methods. The results reported for MAAS-TAN are obtained from a two-stream model composed of 13 temporally linked local graphs, which span about 1.59 seconds. We set $K=3$ for the number of nearest neighbors in the dynamic stream and limit the number of video nodes to 4 per frame. The results reported for  MAAS-LAN are obtained from a two-stream model, which includes a single timestamp and 4 video nodes. For sequences with 5 or more visible speakers, we make sure that one video node contains the features from the active speaker, and randomly sample the remaining three. If no active speaker is present, we just randomly sample 4 speakers without replacement. At inference time, we split the speakers in non overlapping groups of 4, and perform multiple forward passes. Results in the validation set are summarized in Table \ref{tab:sota}.

Our best model, MAAS-TAN, ranks first on the AVA-ActiveSpeaker validation set. We highlight two aspects of these results. First, at 88.8\% mAP, MAAS-TAN outperforms the best results reported on this dataset by at least 1.7\%. It must be noted that some state-of-the-art methods \cite{chung2019naver, zhangmulti} rely on large 3D models and large-scale audiovisual pre-training, while MAAS uses only the standard ImageNet initialization for both streams. Second, while the MAAS-LAN network does not achieve state-of-the-art performance, it outperforms every other method that does not rely on long-term temporal processing \cite{roth2019ava, zhangmulti}. It also remains competitive with those methods that rely only on long-term context \cite{chung2019naver, alcazar2020active}, being outperformed only by the temporal version of \cite{alcazar2020active} by a margin of 0.6\% and falling 2.1\% behind the full method of \cite{alcazar2020active} (temporal context and multi-speaker).

After evaluating the performance of our MAAS method against state-of-the-art techniques, we ablate our best model  (MAAS-TAN) to validate the individual contributions of each design choice, namely: network depth, network width, independent stream contributions, and the number of neighbors for the dynamic stream.
\begin{table}[t]
    \small
    \centering
    \begin{tabular}{c c c c c c c} 
    \toprule
    \multicolumn{1}{c}{} & \multicolumn{5}{c}{ \textbf{Per LAN Video Nodes}} \\
    \multicolumn{1}{c}{\textbf{Number of LANs}} & $1$ & $2$ & $3$ & $4$ & $5$\\ 
    \midrule
    $1$          & $80.2$ & $84.3$ & $84.9$ & $85.1$ & $85$   \\
    $5$          & $85.4$ & $87.1$ & $87.3$ & $87.4$ & $87.3$  \\ 
    $9$          & $86.6$ & $87.8$ & $87.9$ & $88.3$ & $88.5$     \\ 
    $13$         & $87.1$ & $88.1$ & $88.4$ & \textbf{88.8} & $88.5$      \\  
    $15$         & $87.1$ & $87.9$ & $88.2$ & $88.5$ & $88.4$      \\ 
    \toprule
    \end{tabular}
    \caption{\textbf{ Graph Structure in MAAS.} We ablate the size of the MAAS-TAN network which is the core data structure of our approach. We empirically find it beneficial to model multiple speakers at the same time, and find the optimal number of speakers to be 4. Likewise, longer temporal sampling favors the performance but diminishes with sequences longer than 13 frames. }
    \label{tab:context_size}
\end{table}

\paragraph{Network Architecture} We begin by ablating the proposed architecture. We explore the effects of changing the network depth, layer size, and the number of neighbors ($K$) for the dynamic stream. We also control for the individual contribution of each stream.

\begin{table*}[t]
\small
\centering
\begin{tabular}{r c }
\hline
\textbf{Network Depth}  & \textbf{mAP} \\
\hline 
1 Layer & 88.0 \\
2 Layer & 88.2 \\
3 Layer & 88.4 \\
4 Layer & \textbf{88.8} \\
5 Layer & 87.5 \\
\hline
\multicolumn{2}{c}{ \textbf{a)} mAP by Network Depth}
\end{tabular}
\quad
\quad
\begin{tabular}{r c }
\hline
\textbf{Filters in Layer}  & \textbf{mAP} \\
\hline 
32 & 88.5 \\
64 & \textbf{88.8} \\
128 & 88.6\\
256 & 88.1\\
\hline
\multicolumn{2}{c}{\textbf{b)} mAP by Network Width} \\
& \\
\end{tabular}
\quad
\quad
\begin{tabular}{c c c}
\hline
\textbf{Dynamic}  & \textbf{Static}  &  \\
\textbf{Graph}  & \textbf{Graph}  & \textbf{mAP} \\
\hline 
\cmark & \xmark  & 66.5 \\
\xmark & \cmark & 87.9 \\
\cmark & \cmark & \textbf{88.8}\\
\hline
\multicolumn{3}{c}{\textbf{c)} mAP by Individual Stream} \\
& & \\
\end{tabular}
\quad
\quad
\begin{tabular}{r c }
\hline
\textbf{nNighbors}  & \textbf{mAP} \\
\hline 
2 & 88.5\\
3 & \textbf{88.8}\\
4 & 88.4\\
5 & 88.4\\
\hline
\multicolumn{2}{c}{\textbf{d)} mAP by Neighbors } \\
& \\
\end{tabular}
\\

\caption{\textbf{Architecture choices in MAAS.} We ablate the design choices in our proposed GCN-based MAAS-TAN network. We analyse the network depth in \textbf{a)}, and empirically find that a deeper network favors the final result, but saturates at 4 layers. We also analyse the number of filters per layer in \textbf{b)} and find the optimal to be at 64. From \textbf{c)}, we observe that the static stream is far more effective by itself than the dynamic stream; however, the latter stream still incorporates information that is complementary leading to overall improvement. In \textbf{d)}, we empirically find the most suitable number of neighbors in the dynamic stream and set it to 3.}
    
\label{tab:lots}
 \vspace{-0.25cm}
\end{table*}

We summarize our ablation results for the MAAS-TAN network in Table \ref{tab:lots}. In \ref{tab:lots}-a), we identify the depth of the network as a relevant hyper-parameter for its performance. Shallow networks underperform, but increasingly get better as depth increases, reaching an optimal value at 4 layers. Deeper networks have a better capacity for estimating useful features and have the chance to propagate relevant features over a large number of connected nodes, not only the immediate neighbors. In \ref{tab:lots}-b), we show that wider networks have a beneficial effect but saturate quickly with 64 or more filters. Beyond that size, the networks do not yield improvements at the expense of additional network complexity. In \ref{tab:lots}-c), we demonstrate the complementary nature of the two stream approach in MAAS. While the static stream has the best individual performance, the dynamic stream is capable of finding relationships that are beyond the insights we use to create the static graph structure, thus increasing the final performance by 0.9\%. Finally, \ref{tab:lots}-d) shows how the selected number of clusters on the dynamic stream affects the final performance of MAAS. Interestingly, the optimal number of neighbors ($K=3$) matches the number of valid assignations in the active speaker problem (audio with speech, active speaker), (audio with speech, silent speaker) and (audio with silence, silent speaker).

\paragraph{Graph Structure} After assessing the design choices in the architecture, we proceed to evaluate the proposed graph structure. Here, we test for the incremental addition of LAN graphs into a TAN graph that analyses $N$ timestamps. Additionally, we test for the maximum number of video nodes that get linked to an audio node at training time. Table \ref{tab:context_size} summarizes these results. Overall, we notice that MAAS benefits from  modelling longer temporal sequences or modelling more visible speakers. We interpret this as a consequence of our modelling strategy that focuses on the assignation of locally consistent visual and audio patterns, while remaining compatible with the mainstream approach of modelling long-term temporal sequences.

\subsection{Dataset Properties}
We continue our analysis following the evaluation protocol of \cite{roth2019ava} and report MAAS-TAN results in known hard scenarios, namely  multiple possible speakers and small faces.

In Table \ref{tab:scenarios}, we provide a breakdown of MAAS results according to the number of possible speakers. Overall, we see a significant performance increase when comparing MAAS to the AVA baseline \cite{roth2019ava} and improvements in all scenarios when compared to the multi-speaker stack of  \cite{alcazar2020active}. Clearly, the multi-speaker scenario is still quite challenging, but the improvements highlight that our speech assignation-based method is especially effective when two or more possible speakers are present. 

\begin{table}[h]
    \small
    \centering
    \begin{tabular}{c c c c }
        \hline
        \textbf{Number of Faces}  & \textbf{MAAS}  & \textbf{AVA Baseline} \cite{roth2019ava} & \textbf{ASC} \cite{alcazar2020active} \\
        \hline 
        1 & \textbf{93.3} & 87.9 & 91.8 \\
        2 & \textbf{85.8} & 71.6 & 83.8 \\
        3 & \textbf{68.2} & 54.4 & 67.6 \\
        \hline
    \end{tabular}
    
    \caption{\textbf{Performance evaluation by number of faces.} 
    We evaluate MAAS according to the number of  faces visible in the video frame. While performance decreases with more visible people, our method outperforms the AVA baseline and current state-of-the-art.
    }
    \label{tab:scenarios}
\end{table}

In Table \ref{tab:sizes}, we provide a breakdown of MAAS results according to the size of the face crop. We follow the evaluation procedure of \cite{roth2019ava} and create 3 sets of faces: (S) denotes faces smaller than 64$\times$64 pixels, (M) denotes faces between 64$\times$64 and 128$\times$128 pixels, and (L) denotes any face larger than 128$\times$128 pixels. Although MAAS does not explicitly addresses specific face sizes, we observe a large performance gap when compared to the AVA baseline, and we improve in most scenarios when compared to the method of Alcazar \etal \cite{alcazar2020active}. We think this increase in performance is a consequence of better predictions in related faces, \ie smaller faces are typically seen in cluttered scenes with multiple other visible individuals, so our method improves the prediction on these smaller faces by integrating more reliable information from other speakers.

\begin{table}[h]
    \small
    \centering
    \begin{tabular}{c c c c }
        \hline
        \textbf{Face Size}  & \textbf{MAAS}  & \textbf{AVA Baseline} \cite{roth2019ava} & \textbf{ASC} \cite{alcazar2020active} \\
        \hline 
        S & 55.2 & 44.9 & \textbf{56.2} \\
        M & \textbf{79.4} & 68.3 & 79.0 \\
        L & \textbf{93.0} & 86.4 & 92.2 \\
        \hline
    \end{tabular}
    
    \caption{\textbf{Performance evaluation by face size.} 
    We evaluate MAAS in another challenging scenario: small and medium sized faces, which cover less than 128$\times$128 pixels and 64$\times$64 pixels, respectively. We observe that MAAS outperforms the current state-of-the-art, in most scenarios.
    }
    \label{tab:sizes}
\end{table}

To conclude this section we report a final result for MAAS on the AVA-ActiveSpeaker dataset. We assess the relevance of supervision for audio nodes, we empirically find that, by supervising only the video nodes in MASS the performance is slightly reduced to 88.5\%. We think this additional supervision over audio nodes allows for better estimation of the speech events, mitigating some false positives on the speaker detection.

\noindent \textbf{Dubbed audio.} The AVA-ActiveSpeaker dataset contains videos with its native audio as well as dubbed audio. We assess the effect of this imperfect match between the actor's facial gestures and the audio track resulting from dubbing. Since the dataset metadata does not indicate which videos are dubbed, we manually label the validation set videos as either having  native audio (20 videos) or  dubbed audio (13 videos). MAAS achieves an mAP of $86.6\%$ for videos with native audio and $91.6\%$ for dubbed videos. This suggests that MAAS is a viable option for dubbed videos, as it is not so sensitive to the mismatch between facial gestures and the audio waveform resulting from dubbing.

\section{The Talkies Dataset} 
\label{sec:talkies}

Given the scarcity of in-the-wild active speaker datasets, we introduce ``Talkies", a manually labeled dataset for the active speaker detection task. Talkies contains 23,507 face tracks extracted from 421,997 labeled frames that yield a total of 799,446 individual face detections.

In comparison, the Columbia dataset \cite{chakravarty2016active} has about 150,000 face crops, while AVA-ActiveSpeaker \cite{roth2019ava} contains about 5.3 millions (760,000 in validation). Although AVA-ActiveSpeaker has a larger number of individual samples, we argue that Talkies is an interesting, complementary benchmark for three reasons. First, Talkies is more focused on the challenging multi-speaker scenario with 2.3 speakers per frame on average, while AVA-ActiveSpeaker averages only 1.6 speakers per frame. Second, Talkies does not focus on a single source of videos, as in AVA-ActiveSpeaker (Hollywood movies). As a consequence, Talkies contains a more diverse set of actors and scenes, with actors rarely overlapping between clips. This strikes a hard contrast with Hollywood movies, where a small cast takes most of the screen time. Finally, out of screen speech (another challenge for active speaker detection) is not common in Hollywood movies, but it appears more often in Talkies.

\subsection{Creating the Talkies Dataset}
\begin{figure}[h!]
    \begin{center}
        \includegraphics[width=0.45\textwidth]{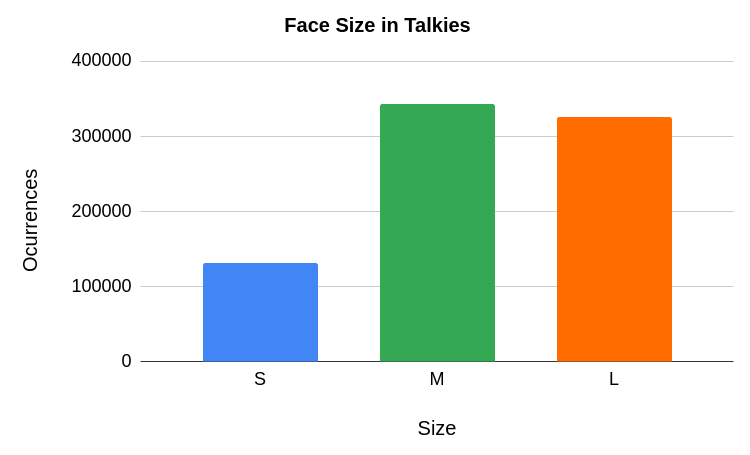} \\
        \scriptsize
        \textbf{a)} Face Size Distribution in Talkies.
        
        \includegraphics[width=0.45\textwidth]{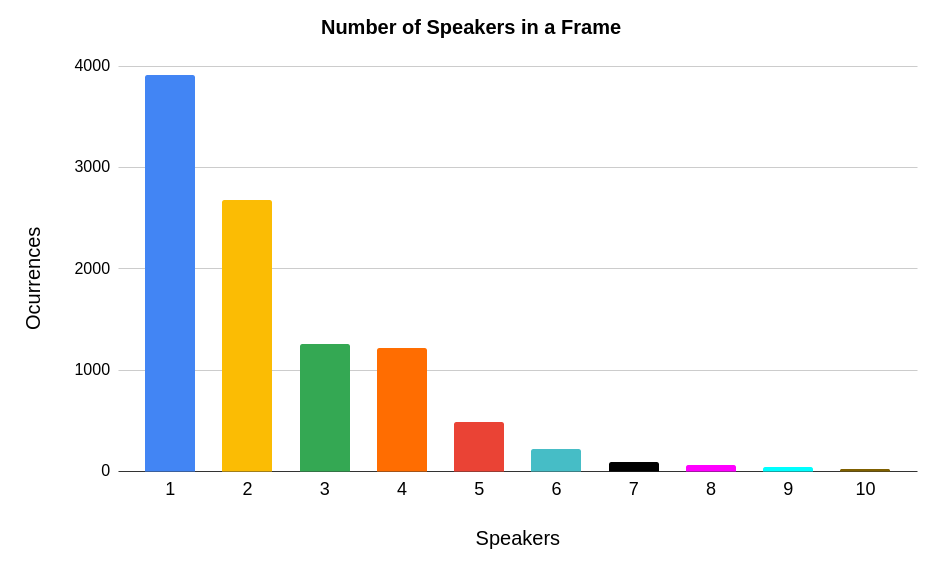} \\
        \scriptsize
        \textbf{b)} Number of Simultaneous Visible Speakers in Talkies.
       
        \includegraphics[width=0.45\textwidth]{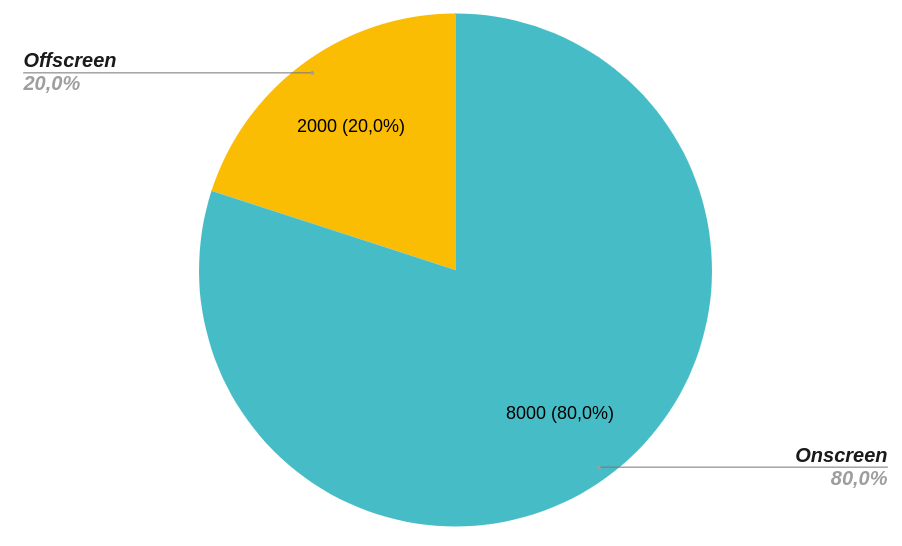} \\
        \scriptsize
        \textbf{c)} Videos With Off-screen Speech labels in Talkies.
    \end{center}
    \caption{\textbf{Talkies Dataset Properties.} We analyze 3 relevant properties in the talkies dataset. In a) we show that most of the dataset contains medium to large faces (64x64 and above), only 16.3\% of Talkies is composed of small faces. In b) we show that most scenes (59.9\%) contain multiple speakers although the most common scenario is a single speaker. In c) we show the proportion of videos labeled with off-screen speech. Talkies is the first active speaker detection dataset to address such scenario.
    }
    \label{fig:talkies_fig}
\end{figure}

We complement the description of Section 5 on the talkies dataset. Our main goal collecting Talkies is  to create a diverse dataset that contains complex scenarios for the active speaker task including: multi-speaker, out of screen speech and noise. Therefore, we use Youtube as our main source of videos, and gather an initial set of clips using search keywords that are likely to contain speech and multiple visible humans, among others: "interview", "daily vlog", "family vlog", "reacting to X", "commentary/review of X". We do not include any further constrain on this initial search.

After gathering an initial pool of videos, we run the voice activity detector of \cite{wan2018generalized} and the face detector of \cite{deng2019retinaface}. This helps us eliminate a large part of the temporal segments, as we keep only a subset where we simultaneously detect human speech and at least 1 face. We do not include any further filtering or manual selection at this step, as it allows for a key element in talkies: the out of screen speech when a set of silent faces is visible and speech from a person out of screen is heard.  On the final step, we want to favor diversity, so we randomly select a single clip per video. This process results in 10.000 short clips (1.5 seconds each) which the video set of Talkies. 

We take the full set of videos in talkies and run the tracker of \cite{wojke2017simple}. We manually label each of the estimated tracklets as 'silent' or 'active speaker'. This process results in a total of 23.508 manually labeled tracks composed of 799.446 individual face detections. Additionally, we provide a video-level annotation for the special case of out of screen speech. That is, every tracklet is labeled as 'silent' and the video itself is associated with label 'OFFSCREEN'.

\begin{table*}[t]
    \small
    \centering
    \begin{tabular}{l r r r}

        \toprule
        & \multicolumn{3}{c}{\textbf{Dataset}} \\
        Stat & AVA \cite{roth2019ava} &  Columbia \cite{chakravarty2016active} & Talkies  \\
        \midrule
        Total Video Hours & \textbf{38.5} & $1.3$ & $4.2$ \\
        Total Speech Hours & \textbf{9.5} & $1.2$ & $3.8$ \\
        Total Tracklets & \textbf{58.171} & N/A & $23.508$ \\
        Total Face Crops & \textbf{5.2M} & $62K$ & $0.8$M \\
        Avg. Speakers per Frame & $1.6$ & $2.4$ & \textbf{2.3} \\
        Unique videos & $280$ & $1$ & \textbf{10.000} \\
        Out of screen labels & \xmark & \xmark & \cmark \\
        \toprule    
    
    \end{tabular}
    \caption{\textbf{Dataset Comparison.}  We compare some relevant statistics of the talkies dataset and the other 2 well know datasets for active speaker detection. While talkies is smaller than the Ava-Active speaker dataset it remains an interesting benchmark given its large video pool, diversity and large number of possible speakers.
    }
    \label{tab:talkies_stats}
\end{table*}

In total, Talkies contains 4 hours and 10 minutes of manually labeled video data. Table \ref{tab:talkies_stats} summarizes some of the most important statistics for Talkies and compares it with the other active speakers datasets. AVA-ActiveSpeakers is clearly a larger dataset regarding number of labeled videos and face crops. Nevertheless talkies, has some relevant properties: it has more speakers per frame (2.3 vs 1.6) and is sampled from a much larger video corpus than AVA-ActiveSpeakers (10.000 vs 218). We think Talkies can be an interesting transfer-dataset for future research or used for approaching the active-speaker problem in a semi-supervised strategy.

We calculate some relevant statistics for the complex scenarios in talkies, namely number of simultaneous faces in a video and size of the face crops, figure \ref{fig:talkies_fig} summarizes these results.  Overall, we find that most images in talkies fall in the Large (larger than $128 \times 128$) and Medium category (between $64 \times 64$ and $128 \times 128$), with only a 16 \% of the face crops into the (harder) category of small images. Additionally, we show the distibution of the number of speakers in a frame, like AVA-ActiveSpeakers the most common scenario is to have a single speaker. However, this represent only the 39.1\% percent of the dataset with the remaining 59.9\% having two or more speakers. In consequence, the average number of speakers stands at 2.3, much higher than the 1.6 of AVA-ActiveSpeakers.

\subsection{Ablation Analysis on Talkies}
We complement the dataset statistics, and assess the effectiveness of MAAS and two baseline methods in some well known challenging scenarios. We follow a similar procedure to \cite{roth2019ava}, and ablate these methods in Talkies according to the number of visible faces and the  size of the face.

Table \ref{tab:talkies_size} shows the ablation results according to the face size. Talkies shows a similar behaviour to AVA-ActiveSpeakers where smaller faces (less than $64 \times 64$) are harder to classify and large faces are easier. We will also highlight the improvement of MAAS in every scenario in comparison to \cite{roth2019ava} and \cite{alcazar2020active}, which is consistent with the results in AVA-ActiveSpeakers.

\begin{table}[t]
    \small
    \centering
      \begin{tabular}{c c c c }
        \hline
        \textbf{Faces Size} & \textbf{MAAS}  & \textbf{ASC}  \cite{alcazar2020active} &  \textbf{AVA Baseline} \cite{roth2019ava} \\
        \hline 
        Small & \textbf{57.0} & $55.4$  & $41.7$ \\
        Mid & \textbf{73.9} & $71.6$ &$64.8$\\
        Large & \textbf{84.8} & $83.9$ & $80.1$ \\
        \hline
    \end{tabular}
    \caption{\textbf{Talkies Face Size}  We evaluate MAAS and some baseline methods in Talkies according to the number size of the face in the video. As observed in Ava-ActiveSpeakers, smaller faces are harder to classify.
    }
    \label{tab:talkies_size}
\end{table}

We continue our analysis by testing the performance of different baseline methods on Talkies, according to the number of visible faces. Table \ref{tab:talkies_scenarios} summarizes these results. Again, these numbers are consistent with those obtained in the AVA-ActiveSpeaker dataset. We highlight that a larger portion of the errors are contained in those clips with 3 or more visible speakers. Once again MAAS outperforms the state-of-the-art in every category except the single speaker baseline, where \cite{alcazar2020active} obtains 0.1\% more. 

\begin{table}[t]
    \small
    \centering
    \begin{tabular}{r c c c }
        \hline
        \textbf{Number of Faces}  & \textbf{MAAS}  & \textbf{AVA Baseline} \cite{roth2019ava} & \textbf{ASC} \cite{alcazar2020active} \\
        \hline 
        1 & 84.5 & 83.0 & \textbf{84.6} \\
        2 & \textbf{81.1} & 72.8 & 77.6 \\
        3 or more & \textbf{71.3} & 63.3 & 70.3\\
        \hline
    \end{tabular}
    
    \caption{\textbf{Performance evaluation by number of faces.} 
    We evaluate MAAS and some baseline methods in Talkies according to the number of  faces visible in the video frame. Performance decreases with more visible people.
    }
    \label{tab:talkies_scenarios}
\end{table}

\begin{table}[h]
    \small
    \centering
    \begin{tabular}{l c c c }
        \hline
        \textbf{Training}  & \textbf{MAAS}  & \textbf{AVA Baseline} \cite{roth2019ava} & \textbf{ASC} \cite{alcazar2020active} \\
        \hline 
        AVA & \textbf{79.1} & 71.5 & 77.4 \\
        AVA augmented & 79.7 & N/A & N/A \\
        \hline
    \end{tabular}
    
    \caption{\textbf{Performance on Talkies.} 
    We evaluate MAAS  performance on the Talkies dataset. Without any fine-tuning on Talkies, MAAS (pre-trained on AVA-ActiveSpeaker) outperforms the baseline by 7.6\% and the state-of-the-art by 1.7\%. A simple augmentation targeting out of screen speech during AVA-ActiveSpeaker training leads to a direct improvement in the challenging scenes of Talkies.
    }
    \label{tab:talkies}
\end{table}

Now, we evaluate the transferability of our MAAS method, trained on AVA-ActiveSpeaker, to the Talkies dataset. No fine-tuning is performed in this case. 

\begin{figure*}[t!]
    \begin{center}
        \includegraphics[width=1\textwidth]{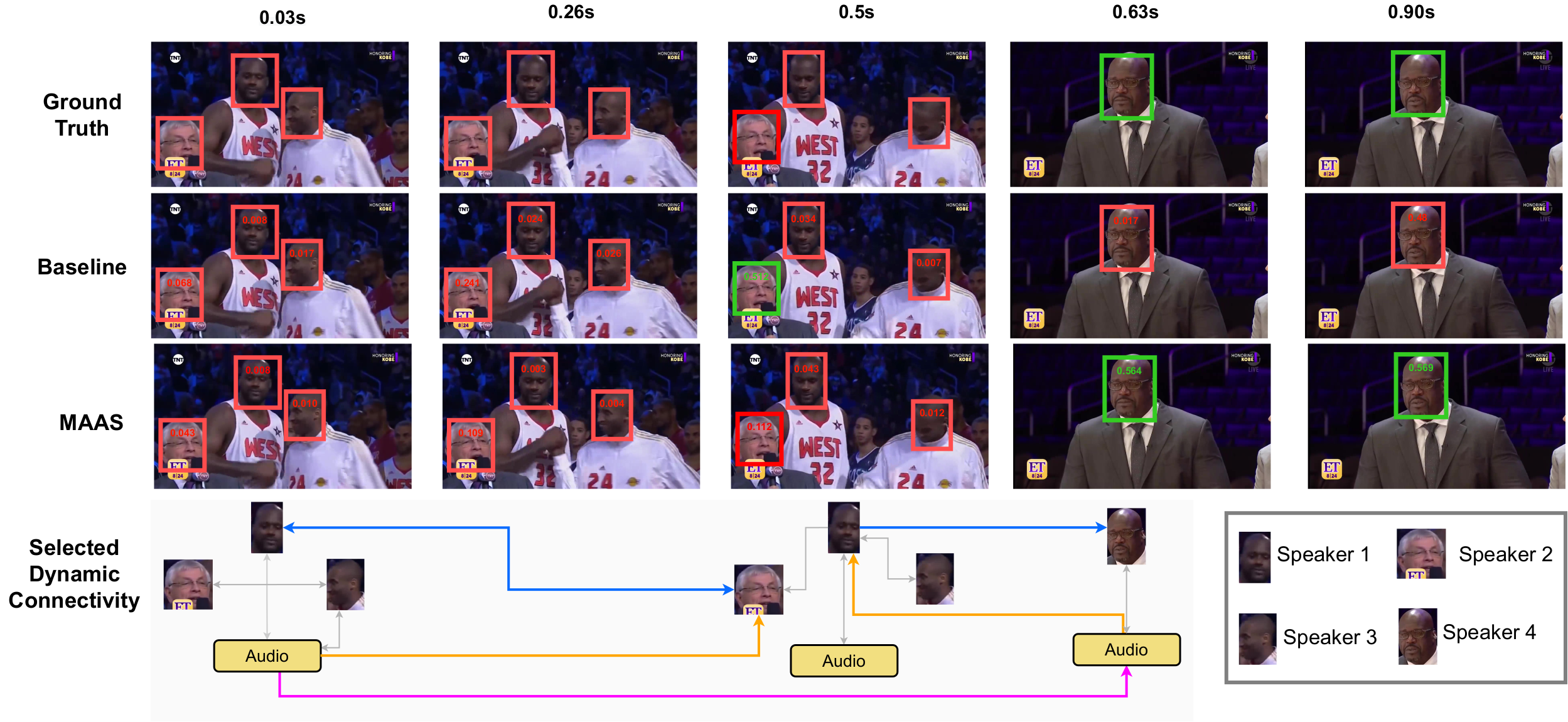}
    \end{center}
    \caption{\textbf{Qualitative Results.} 
    MAAS-TAN includes a dynamic stream that estimates graph structures from nearest neighbours in feature space. The connection patterns in this stream are very diverse and often create edges not present in our static graph. We find that such connectivity allows for information flow between distant audio clips (magenta), inter-speaker relations over multiple time-stamps, and cross-modal arcs that involve nodes in different frames (orange). For easier visualization, we only show a subset of all the dynamic connections.
    }
    \label{fig:qual}
\end{figure*}

In Table \ref{tab:talkies}, we compare the results of  our best model (MAAS-TAN) against the AVA baseline of \cite{roth2019ava} and the ensemble model of \cite{alcazar2020active} on our new dataset. MAAS outperforms these models by 7.6\% and 1.7\%, respectively. These results suggest that the core strategy proposed in MAAS is not domain-specific and can be applied to diverse scenarios without fine-tuning. Moreover, we explore an interesting attribute of Talkies, namely out of screen speech. To do so, we augment the training of MAAS on AVA-ActiveSpeaker, such that we randomly replace a silent audio track (its corresponding frames do not show any active speaker) with a track that contains speech. This simulates out of screen speech scenarios. This artificial substitution is done only during training and with a probability of 20\%. This augmentation does not increase the amount of supervision at training time and has no empirical impact on the performance of MAAS on AVA-ActiveSpeaker, since out of screen speech is not common in Hollywood movies. However, this augmentation strategy does bring an improvement of 0.6\% on Talkies, indicating that MAAS may be flexible enough to handle scenarios more general than those in AVA-ActiveSpeaker.

\subsection{Qualitative Analysis}
\label{subsec:qualy}

We conclude our assessment of MAAS by briefly looking at the connectivity patterns estimated by the dynamic stream. In Figure \ref{fig:qual}, we show a complex clip from the Talkies dataset, in this clip speaker 4 is the only active speaker. In fact, he narrates over the first frames of this clip. This scenario (out of screen speech) makes it very difficult for the baseline to generate accurate predictions on the first frames resulting in some false positive predictions (see speaker 2). MAAS on the other hand performs significantly better, reducing the false positives and effectively detecting the active speaker. 

Empirically, we find that this clip mAP increases from 64\% (baseline) to 97.9\% (MAAS). We think this improvement is explained by two factors. First, MAAS builds more consistent temporal relationships for individual speakers, as its graph structure enforces consistent assignments across the temporal dimension. Second, the dynamic stream allows for unconventional, yet useful connectivity patterns. We show some of these patterns in the bottom row of Figure  \ref{fig:qual}. In blue, we highlight inter-speaker connections across timestamps. These connections are not part of the static graph structure, and can potentially encode semantic relationships between face crops. In magenta, we highlight audio-to-audio connections that go beyond our initial insight of linking adjacent audio clips. We think these connections allow for long-term temporal consistency between audio clips. Such consistency is key to resolve complex scenarios, as is the case with the narration in the selected clip. Finally, we highlight in orange cross-modal connections of nodes at different time-stamps. These connections also differ from those modeled in our static graph, and they reflect semantic similarities in the audiovisual embedding of MAAS.

\section{Conclusion}
We introduced MAAS, a novel multi-modal assignation technique based on graph convolutional networks, for the active speaker detection task. Our method focuses on directly optimizing a graph that simultaneously detects speech events and estimates the best source (active speaker). Additionally, we present Talkies, a novel benchmark with challenging scenarios for active speaker detection, which serves as a challenging transfer dataset for future research.

\textbf{Acknowledgments.} This work was supported by the King Abdullah University of Science and Technology (KAUST) Office of Sponsored Research through the Visual Computing Center (VCC) funding.

{\small
\bibliographystyle{ieee_fullname}
\bibliography{MAAS}

\begin{thebibliography}{10}\itemsep=-1pt

\bibitem{afouras2020self}
Triantafyllos Afouras, Andrew Owens, Joon~Son Chung, and Andrew Zisserman.
\newblock Self-supervised learning of audio-visual objects from video.
\newblock {\em arXiv preprint arXiv:2008.04237}, 2020.

\bibitem{alcazar2020active}
Juan~Leon Alcazar, Fabian Caba, Long Mai, Federico Perazzi, Joon-Young Lee,
  Pablo Arbelaez, and Bernard Ghanem.
\newblock Active speakers in context.
\newblock In {\em Proceedings of the IEEE/CVF Conference on Computer Vision and
  Pattern Recognition}, pages 12465--12474, 2020.

\bibitem{anguera2012speaker}
Xavier Anguera, Simon Bozonnet, Nicholas Evans, Corinne Fredouille, Gerald
  Friedland, and Oriol Vinyals.
\newblock Speaker diarization: A review of recent research.
\newblock {\em IEEE Transactions on Audio, Speech, and Language Processing},
  20(2):356--370, 2012.

\bibitem{chakravarty2015s}
Punarjay Chakravarty, Sayeh Mirzaei, Tinne Tuytelaars, and Hugo Van~hamme.
\newblock Who's speaking? audio-supervised classification of active speakers in
  video.
\newblock In {\em International Conference on Multimodal Interaction (ICMI)},
  2015.

\bibitem{chakravarty2016active}
Punarjay Chakravarty, Jeroen Zegers, Tinne Tuytelaars, et~al.
\newblock Active speaker detection with audio-visual co-training.
\newblock In {\em International Conference on Multimodal Interaction (ICMI)},
  2016.

\bibitem{chung2019naver}
Joon~Son Chung.
\newblock Naver at activitynet challenge 2019--task b active speaker detection
  (ava).
\newblock {\em arXiv preprint arXiv:1906.10555}, 2019.

\bibitem{chung2020spot}
Joon~Son Chung, Jaesung Huh, Arsha Nagrani, Triantafyllos Afouras, and Andrew
  Zisserman.
\newblock Spot the conversation: speaker diarisation in the wild.
\newblock {\em arXiv preprint arXiv:2007.01216}, 2020.

\bibitem{chung2018voxceleb2}
Joon~Son Chung, Arsha Nagrani, and Andrew Zisserman.
\newblock Voxceleb2: Deep speaker recognition.
\newblock {\em arXiv preprint arXiv:1806.05622}, 2018.

\bibitem{chung2017lip}
Joon~Son Chung, Andrew Senior, Oriol Vinyals, and Andrew Zisserman.
\newblock Lip reading sentences in the wild.
\newblock In {\em CVPR}, 2017.

\bibitem{chung2016out}
Joon~Son Chung and Andrew Zisserman.
\newblock Out of time: automated lip sync in the wild.
\newblock In {\em ACCV}, 2016.

\bibitem{chung2019perfect}
Soo-Whan Chung, Joon~Son Chung, and Hong-Goo Kang.
\newblock Perfect match: Improved cross-modal embeddings for audio-visual
  synchronisation.
\newblock In {\em IEEE International Conference on Acoustics, Speech and Signal
  Processing (ICASSP)}, 2019.

\bibitem{cutler2000look}
Ross Cutler and Larry Davis.
\newblock Look who's talking: Speaker detection using video and audio
  correlation.
\newblock In {\em International Conference on Multimedia and Expo}, 2000.

\bibitem{deng2009imagenet}
Jia Deng, Wei Dong, Richard Socher, Li-Jia Li, Kai Li, and Li Fei-Fei.
\newblock Imagenet: A large-scale hierarchical image database.
\newblock In {\em CVPR}, 2009.

\bibitem{deng2019retinaface}
Jiankang Deng, Jia Guo, Yuxiang Zhou, Jinke Yu, Irene Kotsia, and Stefanos
  Zafeiriou.
\newblock Retinaface: Single-stage dense face localisation in the wild.
\newblock {\em arXiv preprint arXiv:1905.00641}, 2019.

\bibitem{everingham2009taking}
Mark Everingham, Josef Sivic, and Andrew Zisserman.
\newblock Taking the bite out of automated naming of characters in tv video.
\newblock {\em Image and Vision Computing}, 27(5):545--559, 2009.

\bibitem{Fey/Lenssen/2019}
Matthias Fey and Jan~E. Lenssen.
\newblock Fast graph representation learning with {PyTorch Geometric}.
\newblock In {\em ICLR Workshop on Representation Learning on Graphs and
  Manifolds}, 2019.

\bibitem{garcia2017speaker}
Daniel Garcia-Romero, David Snyder, Gregory Sell, Daniel Povey, and Alan
  McCree.
\newblock Speaker diarization using deep neural network embeddings.
\newblock In {\em 2017 IEEE International Conference on Acoustics, Speech and
  Signal Processing (ICASSP)}, pages 4930--4934. IEEE, 2017.

\bibitem{Gkioxari2019Mesh}
Georgia Gkioxari, Jitendra Malik, and Justin Johnson.
\newblock Mesh r-cnn.
\newblock {\em arXiv preprint arXiv:1906.02739}, 2019.

\bibitem{hadsell2006dimensionality}
Raia Hadsell, Sumit Chopra, and Yann LeCun.
\newblock Dimensionality reduction by learning an invariant mapping.
\newblock In {\em CVPR}, 2006.

\bibitem{he2016deep}
Kaiming He, Xiangyu Zhang, Shaoqing Ren, and Jian Sun.
\newblock Deep residual learning for image recognition.
\newblock In {\em CVPR}, 2016.

\bibitem{he2016identity}
Kaiming He, Xiangyu Zhang, Shaoqing Ren, and Jian Sun.
\newblock Identity mappings in deep residual networks.
\newblock In {\em European conference on computer vision}, pages 630--645.
  Springer, 2016.

\bibitem{cv_action_jain2016structural}
Ashesh Jain, Amir~R Zamir, Silvio Savarese, and Ashutosh Saxena.
\newblock Structural-rnn: Deep learning on spatio-temporal graphs.
\newblock In {\em Proceedings of the IEEE Conference on Computer Vision and
  Pattern Recognition}, pages 5308--5317, 2016.

\bibitem{jati2019neural}
Arindam Jati and Panayiotis Georgiou.
\newblock Neural predictive coding using convolutional neural networks toward
  unsupervised learning of speaker characteristics.
\newblock {\em IEEE/ACM Transactions on Audio, Speech, and Language
  Processing}, 27(10):1577--1589, 2019.

\bibitem{cv_inv_scene_johnson2018image}
Justin Johnson, Agrim Gupta, and Li Fei-Fei.
\newblock Image generation from scene graphs.
\newblock In {\em Proceedings of the IEEE Conference on Computer Vision and
  Pattern Recognition}, pages 1219--1228, 2018.

\bibitem{karpathy2014large}
Andrej Karpathy, George Toderici, Sanketh Shetty, Thomas Leung, Rahul
  Sukthankar, and Li Fei-Fei.
\newblock Large-scale video classification with convolutional neural networks.
\newblock In {\em Proceedings of the IEEE conference on Computer Vision and
  Pattern Recognition}, pages 1725--1732, 2014.

\bibitem{kim2018learning}
Changil Kim, Hijung~Valentina Shin, Tae-Hyun Oh, Alexandre Kaspar, Mohamed
  Elgharib, and Wojciech Matusik.
\newblock On learning associations of faces and voices.
\newblock In {\em ACCV}, 2018.

\bibitem{kingma2014adam}
D Kinga and J~Ba Adam.
\newblock A method for stochastic optimization.
\newblock In {\em ICLR}, 2015.

\bibitem{kipf2016semi}
Thomas~N Kipf and Max Welling.
\newblock Semi-supervised classification with graph convolutional networks.
\newblock {\em arXiv preprint arXiv:1609.02907}, 2016.

\bibitem{Li_2019_ICCV}
Guohao Li, Matthias Muller, Ali Thabet, and Bernard Ghanem.
\newblock Deepgcns: Can gcns go as deep as cnns?
\newblock In {\em The IEEE International Conference on Computer Vision (ICCV)},
  2019.

\bibitem{li2019deepgcns_journal}
Guohao Li, Matthias Müller, Guocheng Qian, Itzel~C. Delgadillo, Abdulellah
  Abualshour, Ali Thabet, and Bernard Ghanem.
\newblock Deepgcns: Making gcns go as deep as cnns, 2019.

\bibitem{li2019sgas}
Guohao Li, Guocheng Qian, Itzel~C. Delgadillo, Matthias Müller, Ali Thabet,
  and Bernard Ghanem.
\newblock Sgas: Sequential greedy architecture search, 2019.

\bibitem{li2020deepergcn}
Guohao Li, Chenxin Xiong, Ali Thabet, and Bernard Ghanem.
\newblock Deepergcn: All you need to train deeper gcns.
\newblock {\em arXiv preprint arXiv:2006.07739}, 2020.

\bibitem{cv_scene_li2018factorizable}
Yikang Li, Wanli Ouyang, Bolei Zhou, Jianping Shi, Chao Zhang, and Xiaogang
  Wang.
\newblock Factorizable net: an efficient subgraph-based framework for scene
  graph generation.
\newblock In {\em Proceedings of the European Conference on Computer Vision
  (ECCV)}, pages 335--351, 2018.

\bibitem{nagrani2018learnable}
Arsha Nagrani, Samuel Albanie, and Andrew Zisserman.
\newblock Learnable pins: Cross-modal embeddings for person identity.
\newblock In {\em ECCV}, 2018.

\bibitem{nagrani2018seeing}
Arsha Nagrani, Samuel Albanie, and Andrew Zisserman.
\newblock Seeing voices and hearing faces: Cross-modal biometric matching.
\newblock In {\em CVPR}, 2018.

\bibitem{nagrani2017voxceleb}
Arsha Nagrani, Joon~Son Chung, and Andrew Zisserman.
\newblock Voxceleb: a large-scale speaker identification dataset.
\newblock {\em arXiv preprint arXiv:1706.08612}, 2017.

\bibitem{ngiam2011multimodal}
Jiquan Ngiam, Aditya Khosla, Mingyu Kim, Juhan Nam, Honglak Lee, and Andrew~Y
  Ng.
\newblock Multimodal deep learning.
\newblock In {\em ICML}, 2011.

\bibitem{owens2018audio}
Andrew Owens and Alexei~A Efros.
\newblock Audio-visual scene analysis with self-supervised multisensory
  features.
\newblock In {\em Proceedings of the European Conference on Computer Vision
  (ECCV)}, pages 631--648, 2018.

\bibitem{paszke2017automatic}
Adam Paszke, Sam Gross, Soumith Chintala, Gregory Chanan, Edward Yang, Zachary
  DeVito, Zeming Lin, Alban Desmaison, Luca Antiga, and Adam Lerer.
\newblock Automatic differentiation in pytorch.
\newblock In {\em NeurIPS-Workshop}, 2017.

\bibitem{qi20173d}
Xiaojuan Qi, Renjie Liao, Jiaya Jia, Sanja Fidler, and Raquel Urtasun.
\newblock 3d graph neural networks for rgbd semantic segmentation.
\newblock In {\em Proceedings of the IEEE International Conference on Computer
  Vision}, pages 5199--5208, 2017.

\bibitem{ravanelli2018speaker}
Mirco Ravanelli and Yoshua Bengio.
\newblock Speaker recognition from raw waveform with sincnet.
\newblock In {\em IEEE Spoken Language Technology Workshop (SLT)}, 2018.

\bibitem{roth2019ava}
Joseph Roth, Sourish Chaudhuri, Ondrej Klejch, Radhika Marvin, Andrew
  Gallagher, Liat Kaver, Sharadh Ramaswamy, Arkadiusz Stopczynski, Cordelia
  Schmid, Zhonghua Xi, et~al.
\newblock Ava-activespeaker: An audio-visual dataset for active speaker
  detection.
\newblock {\em arXiv preprint arXiv:1901.01342}, 2019.

\bibitem{saenko2005visualspeech}
Kate Saenko, Karen Livescu, Michael Siracusa, Kevin Wilson, James Glass, and
  Trevor Darrell.
\newblock Visual speech recognition with loosely synchronized feature streams.
\newblock In {\em ICCV}, 2005.

\bibitem{sharma2020crossmodal}
Rahul Sharma, Krishna Somandepalli, and Shrikanth Narayanan.
\newblock Crossmodal learning for audio-visual speech event localization.
\newblock {\em arXiv preprint arXiv:2003.04358}, 2020.

\bibitem{shum2013unsupervised}
Stephen~H Shum, Najim Dehak, R{\'e}da Dehak, and James~R Glass.
\newblock Unsupervised methods for speaker diarization: An integrated and
  iterative approach.
\newblock {\em IEEE Transactions on Audio, Speech, and Language Processing},
  21(10):2015--2028, 2013.

\bibitem{tao2017bimodal}
Fei Tao and Carlos Busso.
\newblock Bimodal recurrent neural network for audiovisual voice activity
  detection.
\newblock In {\em INTERSPEECH}, pages 1938--1942, 2017.

\bibitem{tranter2006overview}
Sue~E Tranter and Douglas~A Reynolds.
\newblock An overview of automatic speaker diarization systems.
\newblock {\em IEEE Transactions on audio, speech, and language processing},
  14(5):1557--1565, 2006.

\bibitem{waibel1989phoneme}
Alex Waibel, Toshiyuki Hanazawa, Geoffrey Hinton, Kiyohiro Shikano, and Kevin~J
  Lang.
\newblock Phoneme recognition using time-delay neural networks.
\newblock {\em IEEE transactions on acoustics, speech, and signal processing},
  37(3):328--339, 1989.

\bibitem{wan2018generalized}
Li Wan, Quan Wang, Alan Papir, and Ignacio~Lopez Moreno.
\newblock Generalized end-to-end loss for speaker verification.
\newblock In {\em 2018 IEEE International Conference on Acoustics, Speech and
  Signal Processing (ICASSP)}, pages 4879--4883. IEEE, 2018.

\bibitem{wang2018speaker}
Quan Wang, Carlton Downey, Li Wan, Philip~Andrew Mansfield, and Ignacio~Lopz
  Moreno.
\newblock Speaker diarization with lstm.
\newblock In {\em 2018 IEEE International Conference on Acoustics, Speech and
  Signal Processing (ICASSP)}, pages 5239--5243. IEEE, 2018.

\bibitem{wang2018voicefilter}
Quan Wang, Hannah Muckenhirn, Kevin Wilson, Prashant Sridhar, Zelin Wu, John
  Hershey, Rif~A Saurous, Ron~J Weiss, Ye Jia, and Ignacio~Lopez Moreno.
\newblock Voicefilter: Targeted voice separation by speaker-conditioned
  spectrogram masking.
\newblock {\em arXiv preprint arXiv:1810.04826}, 2018.

\bibitem{wang2018dynamic}
Yue Wang, Yongbin Sun, Ziwei Liu, Sanjay Sarma, Michael Bronstein, and Justin
  Solomon.
\newblock Dynamic graph cnn for learning on point clouds.
\newblock {\em ACM Transactions on Graphics}, 2018.

\bibitem{wang2019dynamic}
Yue Wang, Yongbin Sun, Ziwei Liu, Sanjay~E Sarma, Michael~M Bronstein, and
  Justin~M Solomon.
\newblock Dynamic graph cnn for learning on point clouds.
\newblock {\em Acm Transactions On Graphics (tog)}, 38(5):1--12, 2019.

\bibitem{wojke2017simple}
Nicolai Wojke, Alex Bewley, and Dietrich Paulus.
\newblock Simple online and realtime tracking with a deep association metric.
\newblock In {\em 2017 IEEE international conference on image processing
  (ICIP)}, pages 3645--3649. IEEE, 2017.

\bibitem{xie2019clouds}
Zhuyang Xie, Junzhou Chen, and Bo Peng.
\newblock Point clouds learning with attention-based graph convolution
  networks.
\newblock {\em arXiv preprint arXiv:1905.13445}, 2019.

\bibitem{cv_scene_xu2017scene}
Danfei Xu, Yuke Zhu, Christopher~B Choy, and Li Fei-Fei.
\newblock Scene graph generation by iterative message passing.
\newblock In {\em Proceedings of the IEEE Conference on Computer Vision and
  Pattern Recognition}, pages 5410--5419, 2017.

\bibitem{g_tad}
Mengmeng Xu, Chen Zhao, David~S Rojas, Ali Thabet, and Bernard Ghanem.
\newblock G-tad: Sub-graph localization for temporal action detection.
\newblock In {\em Proceedings of the IEEE/CVF Conference on Computer Vision and
  Pattern Recognition}, pages 10156--10165, 2020.

\bibitem{yadav2018learning}
Sarthak Yadav and Atul Rai.
\newblock Learning discriminative features for speaker identification and
  verification.
\newblock In {\em Interspeech}, 2018.

\bibitem{cv_action_yan2018spatial}
Sijie Yan, Yuanjun Xiong, and Dahua Lin.
\newblock Spatial temporal graph convolutional networks for skeleton-based
  action recognition.
\newblock In {\em Thirty-Second AAAI Conference on Artificial Intelligence},
  2018.

\bibitem{cv_scene_yang2018graph}
Jianwei Yang, Jiasen Lu, Stefan Lee, Dhruv Batra, and Devi Parikh.
\newblock Graph r-cnn for scene graph generation.
\newblock In {\em Proceedings of the European Conference on Computer Vision
  (ECCV)}, pages 670--685, 2018.

\bibitem{yu2017active}
Chengzhu Yu and John~HL Hansen.
\newblock Active learning based constrained clustering for speaker diarization.
\newblock {\em IEEE/ACM Transactions on Audio, Speech, and Language
  Processing}, 25(11):2188--2198, 2017.

\bibitem{zhang2019fully}
Aonan Zhang, Quan Wang, Zhenyao Zhu, John Paisley, and Chong Wang.
\newblock Fully supervised speaker diarization.
\newblock In {\em IEEE International Conference on Acoustics, Speech and Signal
  Processing (ICASSP)}. IEEE, 2019.

\bibitem{zhangmulti}
Yuan-Hang Zhang, Jingyun Xiao, Shuang Yang, and Shiguang Shan.
\newblock Multi-task learning for audio-visual active speaker detection.

\end{thebibliography}
}

\end{document}